\documentclass[conference]{IEEEtran}  

\IEEEoverridecommandlockouts                  

\usepackage{times}
\usepackage{enumitem}
\usepackage{amsmath}
\usepackage{amsfonts}
\usepackage{graphicx}
\usepackage{float} 
\usepackage{booktabs}
\usepackage{multirow}
\usepackage{multicol}
\usepackage{caption}
\usepackage{algorithm}
\usepackage{algpseudocode}
\usepackage{multicol}
\usepackage[numbers,sort&compress,square]{natbib}
\usepackage[bookmarks=true]{hyperref}
\hypersetup{
    colorlinks=true,
    linkcolor=blue,
    filecolor=blue,      
    urlcolor=blue,
    citecolor=magenta,
}
\usepackage{xcolor}

\DeclareMathOperator*{\argmin}{arg\,min}
\DeclareMathOperator*{\argmax}{arg\,max}
\title{\LARGE \bf
XMoP: Whole-Body Control Policy for Zero-shot Cross-Embodiment Neural Motion Planning
}

\author{Prabin Kumar Rath and Nakul Gopalan \\ Arizona State University
\thanks{Videos and code are available at \href{https://prabinrath.github.io/xmop}{https://prabinrath.github.io/xmop}.}
}

\makeatletter
\g@addto@macro\@maketitle{
  \vspace*{-10pt}
  \begin{figure}[H]
  \setlength{\linewidth}{\textwidth}
  \setlength{\hsize}{\textwidth}
  \centering
  \includegraphics[width=\textwidth]{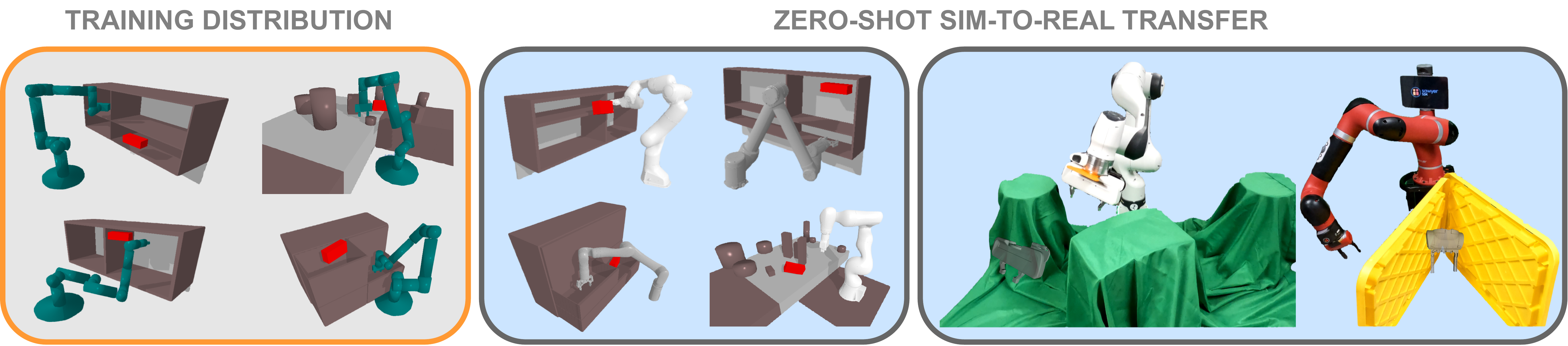}
  \caption{\small XMoP learns configuration-space planning for a distribution of synthetic embodiments (\textit{left}) and \textit{zero-shot} transfers the planning behavior to unseen simulated (\textit{center}) and real-world manipulators (\textit{right}).}
  \label{fig:poster}
  \vspace*{-20pt}
  \end{figure}
  \setcounter{figure}{0}
}
\makeatother

\begin{document}

\maketitle
\setcounter{figure}{1}

\begin{abstract}
Classical manipulator motion planners work across different robot embodiments~\cite{zhou2022review}. However they plan on a pre-specified static environment representation, and are not scalable to unseen dynamic environments. Neural Motion Planners (NMPs) \cite{qureshi2019motion} are an appealing alternative to conventional planners as they incorporate different environmental constraints to learn motion policies directly from raw sensor observations. Contemporary state-of-the-art NMPs can successfully plan across different environments~\cite{fishman2023motion}.  However none of the existing NMPs generalize across robot embodiments. In this paper we propose Cross-Embodiment Motion Policy (XMoP), a neural policy for learning to plan over a distribution of manipulators. XMoP implicitly learns to satisfy kinematic constraints for a distribution of robots and \textit{zero-shot} transfers the planning behavior to unseen robotic manipulators within this distribution. We achieve this generalization by formulating a whole-body control policy that is trained on planning demonstrations from over three million procedurally sampled robotic manipulators in different simulated environments. Despite being completely trained on synthetic embodiments and environments, our policy exhibits strong sim-to-real generalization across manipulators with different kinematic variations and degrees of freedom with a single set of frozen policy parameters. We evaluate XMoP on $7$ commercial manipulators and show successful cross-embodiment motion planning, achieving an average $70\%$ success rate on baseline benchmarks. Furthermore, we demonstrate our policy sim-to-real on two unseen manipulators solving novel planning problems across three real-world domains even with dynamic obstacles.
\end{abstract}

\thispagestyle{empty}
\pagestyle{empty}
\vspace*{-5pt}
\section{Introduction}
Motion planning for robotic manipulators is the task of finding a sequence of robot configurations connecting a start joint state to the goal joint state while respecting joint limits of the robot and avoiding obstacles. Even after decades of research in this domain, real-time motion planning in complex unseen environments is still a challenging problem~\cite{tang2012review, petrovic2018motion, sandakalum2022motion}.

Classical motion planners either use random sampling to explore the configuration-space (C-space) \cite{lavalle2001randomized, kuffner2000rrt, karaman2011sampling, strub2022adaptively} or employ gradient-based optimization methods \cite{ratliff2009chomp, kalakrishnan2011stomp, schulman2013finding, dong2016motion} to search for a valid plan. While these algorithms generalize across embodiments, they often demand a non real-time computation budget for generating desired motion behaviors in geometrically complex environments \cite{petrovic2018motion, strub2022adaptively}. Furthermore, classical algorithms assume the availability of a pre-computed geometric representation of the robot's workspace for state validation, which hinders their scalability in unseen and dynamic environments. To overcome these limitations, neural planners learn to generate trajectories directly from visual observations \cite{qureshi2019motion, huh2021cost, johnson2023learning, yamada2023leveraging, carvalho2023, fishman2023motion}. However, these policies are individually trained on data from a single manipulator, trading-off the cross-embodiment flexibility offered by classical planners that are agnostic to the robot's morphology.

We identify two fundamental constraints for learning cross-embodiment motion planning. First, different manipulators have varying kinematic properties such as link lengths, as well as diverse morphologies characterized by their degrees of freedom. Each manipulator operates within a particular configuration-subspace bounded by its joint limits. Thus, training a single neural policy to generate actions spanning multiple bounded subspaces renders cross-embodiment C-space policies a challenging task to learn. Second, data for training cross-embodiment policies is difficult to gather as there are only a limited number of embodiments available commercially, which do not fully capture the distribution of possible kinematic variations.

To address the above challenges, we present \textbf{Cross}-Embodiment \textbf{Mo}tion \textbf{P}olicy (\textbf{XMoP}), a family of data-driven methods to learn neural policies for cross-embodiment motion planning. Our contributions are outlined as follows:
\begin{itemize}[leftmargin=*]
    \item Our novel control policy utilizes the robot's physical description (i.e., URDF~\cite{tola2023understanding}) and generates C-space plans for a distribution of $6$- and $7$-DoF manipulators. We demonstrate \textit{zero-shot} generalization of our policy to $7$ unseen commercial manipulators, using a set of frozen policy parameters.
    \item We propose a $3$D semantic segmentation-based model for perceptual cross-embodiment collision detection that achieves a $98\%$ recall and \textit{zero-shot} transfers to different real-world unseen planning environments.
    \item Finally, we combine our control policy with the collision model under a model-predictive framework, achieving an average $70\%$ success rate for motion planning using purely visual inputs with manipulators and environments which were never seen during training.
\end{itemize} 
To the best of our knowledge, XMoP is the first configuration space neural planning policy that \textit{zero-shot} transfers to unseen robotic manipulators. We demonstrate sim-to-real transfer along with success rate evaluations on Franka FR3 and Sawyer robots solving novel planning problems in unseen real-world environments.

\section{Related Work}
\label{sec:related_work}
\textbf{Learning to Plan}: Previous studies have proposed learning deep priors for informed configuration-space sampling in RRTs~\cite{ichter2018learning, qureshi2019motion, johnson2023learning}. Another line of work have investigated latent space representations and optimization for motion planning \cite{huh2021cost, yamada2023leveraging}. Motion planning has also been formulated as sampling from a distribution of valid trajectories using diffusion models \cite{janner2022planning, carvalho2023, saha2023}. Recently, behavior cloning has shown success in learning planning policies from synthetic demonstrations \cite{fishman2023motion, saha2023}. In contrast, Reinforcement Learning (RL) has also been used for training unsupervised motion planning policies \cite{jurgenson2019harnessing, strudel2021learning}. Many of these learning based approaches have been evaluated in toy environments and simulations where the environment was fully observable \cite{ichter2018learning, jurgenson2019harnessing, strudel2021learning, janner2022planning, saha2023}, making it unclear how they can be extended for real-world manipulator motion planning. Few methods have been deployed on real robots and show environment generalization \cite{qureshi2019motion, huh2021cost, johnson2023learning, yamada2023leveraging, carvalho2023, fishman2023motion}; however they require a separate policy to be trained for each embodiment. None of the previous works demonstrate motion planning across embodiments, thus lacking the versatility offered by their classical counterparts. In contrast, our method shows \textit{zero-shot} generalization over a distribution of manipulators enabling cross-embodiment neural motion planning.

\textbf{Cross Embodiment Learning}: Prior works have shown embodiment generalization with RL using robot-environment mix-match strategies \cite{devin2017learning}, vector representation of the hardware \cite{chen2018hardware}, and latent representation for robots \cite{ghadirzadehbayesian}. These methods only work on seen manipulators having fixed DoF whereas our method generalizes across manipulators with different DoF. Recent supervised learning methods showcase policy transfer across robots in task-space using robot specific controllers \cite{yang2023polybot, yang2024pushing, shah2023vint, salhotra2023bridging, chen2024mirage}. As these policies do not generalize in robot configuration-space, they are not suitable for motion planning formulation. There has been a surge of interest in meta policies that are trained on procedurally sampled robot morphologies with kinematics and dynamics randomization \cite{gupta2022metamorph, schubert2023generalist}. These policies have shown \textit{zero-shot} generalization to unseen embodiments in simulation, but have never been deployed on robots in the real world. \cite{feng2023genloco} show \textit{zero-shot} generalization to real robots; however they use a fixed DoF template and do not account for morphology variations. None of the prior methods have demonstrated generalization to commercial robots with morphology and kinematic variations whereas our method shows \textit{zero-shot} sim-to-real generalization across different robotic manipulators in the real world.

\section{Methodology}
\begin{figure*}[!t]
\includegraphics[width=\textwidth]{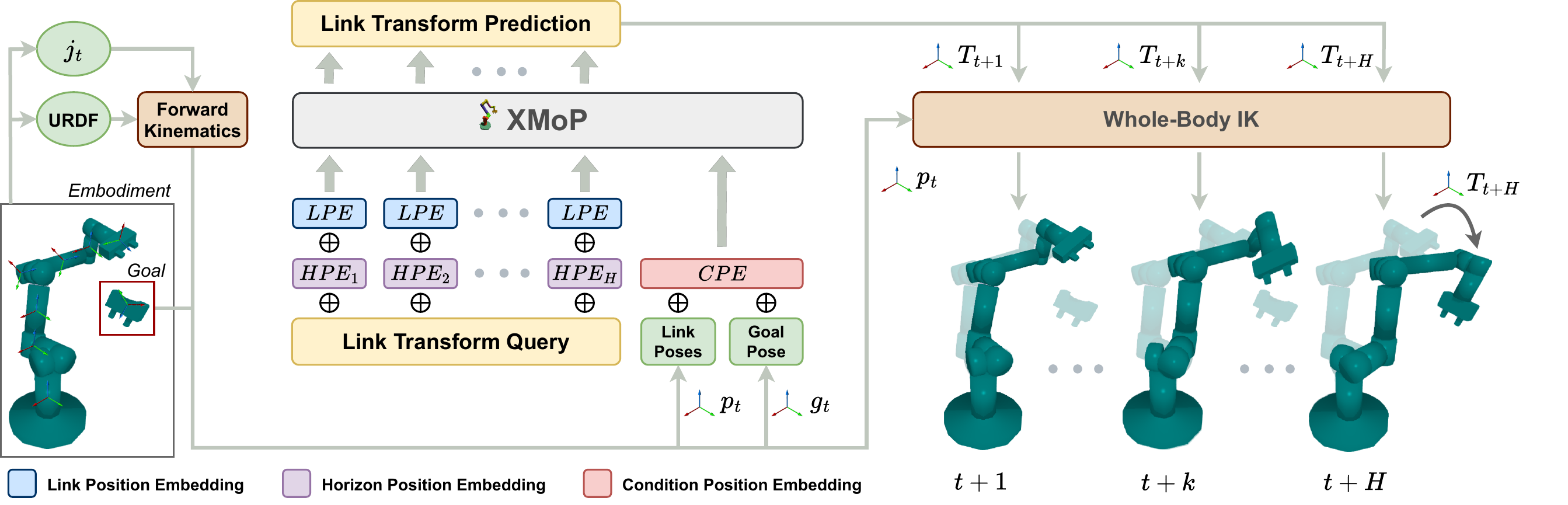}
\centering
\caption{XMoP perceives the embodiment state as a sequence of whole-body SE($3$) link poses $p_t$ and predicts link-wise pose transformations $T_{t+1:t+H}$ over a horizon $H$ to move the end-effector towards the goal pose $g_t$. We use a Transformer~\cite{vaswani2017attention} base policy architecture, that operates on an input pose sequence ($p_t$, $g_t$) and uses self-attention to convert the query tokens into a sequence of link-wise relative pose transformations. Contextual information is provided to the Transformer using three types of position embeddings: (1) LPE is a fixed set of sinusoidal position embeddings that are repeated over the horizon providing kinematic chain awareness for every horizon step; (2) HPE and (3) CPE, are learned position embeddings providing awareness for horizon and input, respectively. Additionally, we use novel attention masking strategies within the Transformer for cross-embodiment adaptation. The predicted link-wise transformations are applied to the instantaneous link poses to reconstruct the future whole-body pose of the manipulator. This target pose is achieved using a whole-body IK procedure, which retrieves the configuration-space joint state within the bounds specified in the manipulator's URDF.
} 
\label{fig:method}
\vspace*{-10pt}
\end{figure*}
\subsection{Whole-Body Control Formulation}
\label{sec:method}
Prior methods for neural motion planning directly predict configuration-space actions that do not generalize across embodiments \cite{qureshi2019motion, huh2021cost, johnson2023learning, yamada2023leveraging, carvalho2023, fishman2023motion}. We formulate XMoP as a Markovian motion dynamics model $f(p_{t+1:t+H}|p_t,g_t)$ that provides the future states of manipulator over a horizon of $H$ steps. The instantaneous state observation for the manipulator is represented as a sequence of rigid-body SE($3$) link poses with respect to the robot's base i.e., $p_t \in \mathbb{R}^{D \times 4 \times 4}$, where $D$ is the number of rigid-body links on the manipulator. The goal $g_t \in \mathbb{R}^{4 \times 4}$ represents the end-effector SE($3$) target for motion planning. Our control policy $\pi(a_t|p_t,g_t)$ as shown in Fig.~\ref{fig:method}, learns to predict link-wise relative SE($3$) transformations $a_t = T_{t+1:t+H}$ for reconstructing the whole-body manipulator pose in future time steps. The formulation for the transformation target $T_{t+k} \in \mathbb{R}^{D \times 4 \times 4}$ for $k \in \{1,...,H\}$ is shown in eq.~\ref{regression_target}.
\begin{equation}
\label{regression_target}
\begin{aligned}
    &T_{t+k} p_t = p_{t+k} \implies T_{t+k} = p_{t+k} (p_{t})^{-1}
\end{aligned}
\end{equation}
where link poses $p_t$ are obtained using the manipulator's forward kinematics function $\phi(j_t)$, with $j_t \in \mathbb{R}^\text{DoF}$ as the instantaneous configuration-space observation. The C-space action in future time step $j_{t+k}$ is retrieved from the predicted whole-body pose $\hat{p}_{t+k}=\hat{T}_{t+k}p_t$ by solving for whole-body IK using the following constrained optimization procedure:
\begin{equation}
\begin{aligned}
    &\min_{j_{t+k}} & & \|\hat{p}_{t+k} - \phi(j_{t+k})\|, \quad \text{s.t.} \quad j_{L} < j_{t+k} < j_{U}
\end{aligned}
\end{equation}
where $j_{L}$ and $j_{U}$ are the lower and upper joint limits of the manipulator. The above optimization objective is non-convex, and hence a close initial guess is required for convergence. We address this issue by collecting dense planning demonstrations with a maximum per-joint deviation of 0.05 rad. Thus, making the instantaneous observation $j_t$ to be an initial guess that lies within the close neighbourhood of $j_{t+k}$. In practice, we also employ multiple retries with random initial joint states to handle redundancy and singularities in manipulators.

\subsection{Pose Transformation Policy}
\label{sec:policy}
We formulate the whole-body control policy $\pi_{\theta}(a_t|p_t,g_t)$ as a stochastic Transformer diffusion policy~\cite{chi2023diffusion} parameterized by $\theta$ that predicts a batch of possible future trajectories for model predictive control. While training, the noise prediction model $\epsilon_\theta$ takes the noisy sample $a^{\tau}_t$, which is obtained by applying the forward diffusion process to $a^{0}_t=T_{t+1:t+H}$, where $\tau$ is the diffusion step. We use the noisy sample $a^{\tau}_t$ as link transform query, which is passed to the diffusion model, as shown in Fig. \ref{fig:method}. Additionally, we also pass $c = (p_t, g_t)$ for observation and goal conditioning. For step conditioning, we follow the adaptive layer normalization strategy proposed in Diffusion Transformers~\cite{peebles2023scalable}. We train the noise prediction model using mean square error loss as shown in eq.~\ref{multistep_loss} which minimizes the variational lower bound of the KL-divergence between the original data distribution and the DDPM~\cite{ho2020denoising} distribution.
\begin{equation}
\label{multistep_loss}
\begin{aligned}
&\mathcal{L}_{xmop}=\|\epsilon_\theta(a^{\tau}_t, c, \tau)-\epsilon\|^{2}_2, \quad \epsilon \in \mathcal{N}(0,I)
\end{aligned}
\end{equation}
We convert the input pose matrices in $p_t$ to compact $9$D representations $r = (\Vec{o},\Vec{x},\Vec{y})$, where $\Vec{o} \in \mathbb{R}^3$ is the $3$D translation component and $(\Vec{x}, \Vec{y}) \in \mathbb{R}^6$ is the $6$D rotation component in SE($3$)~\cite{zhou2019continuity}. The model predicts the relative pose transformations in the same $9$D representation space which is converted back to homogeneous matrices using Gram–Schmidt orthogonal decomposition~\cite{zhou2019continuity}. Our policy utilizes the Transformer~\cite{vaswani2017attention} model as the underlying backbone which expects input in a sequential format. On that aspect, we emphasize on four key design decisions in our policy: 
\begin{enumerate}[leftmargin=5mm, nolistsep]
\item \textbf{SE($3$) Proprioception}: The embodiment state is provided to and queried from the policy as a sequence of SE($3$) \textit{pose-tokens}, allowing the policy to learn motion synergies between rigid-body links.
\item \textbf{Kinematic Masking}: We introduce an inductive bias for kinematics by restricting attention to parent or ancestor links at the current horizon step, and to the same link at both the current and previous horizon steps. 
\item \textbf{Morphology Adaptation}: To enable learning across different morphologies, we mask out the \textit{pose-tokens} for unavailable links. For example, we use $D$ \textit{pose-tokens} per horizon step, with $D=8$ for both $6$- and $7$-DoF robots (including the base link \textit{pose-token} that is perpetually set to identity). For a $6$-DoF robot however, we mask out one \textit{pose-token} and pass $\vec{0}$ to account for the missing link.
\item \textbf{Link-Horizon Position Embedding}: Contextual information is provided to the Transformer using position embeddings for query and input \textit{pose-tokens}. Fig.~\ref{fig:method} shows the position embedding scheme used in XMoP.
\end{enumerate}
Prior works have used link-wise tokens \cite{gupta2022metamorph}, temporal state inputs \cite{janner2022planning, schubert2023generalist}, action diffusion \cite{chi2023diffusion, janner2022planning, carvalho2023}, and SE($3$) pose observations \cite{liu2022structdiffusion, simeonov2023shelving} for learning and planning applications. We combine these ideas with our whole-body pose transformation method to learn neural policy for cross-embodiment motion planning. 

\subsection{Collision-Free Motion Synthesis}
\label{sec:coll_det}
We formulate collision detection as a semantic segmentation problem for identifying the links of the robot that are in collision given a pointcloud observation of the workspace. Our semantic collision detection model \textbf{XCoD} : $\mathbb{R}^4 \rightarrow \mathbb{R}^2$ takes segmented pointcloud of the workspace as input where each point consists of $3$D spatial coordinates, and a semantic label. These points are uniformly sampled from the surface of individual links (URDF mesh files) of the manipulator and scene obstacles for the future time steps as predicted by our control policy $\pi_\theta$. We assign unique semantic labels to each link of the robot, while using a separate reserved label for all obstacles. For training the collision model, we utilize point-wise binary label $y$, where collision-free link points are assigned a training label of $0$, whereas link points in collision are assigned a training label of $1$. 
\begin{figure}[t]
    \centering    
    \includegraphics[width=0.33\textwidth]{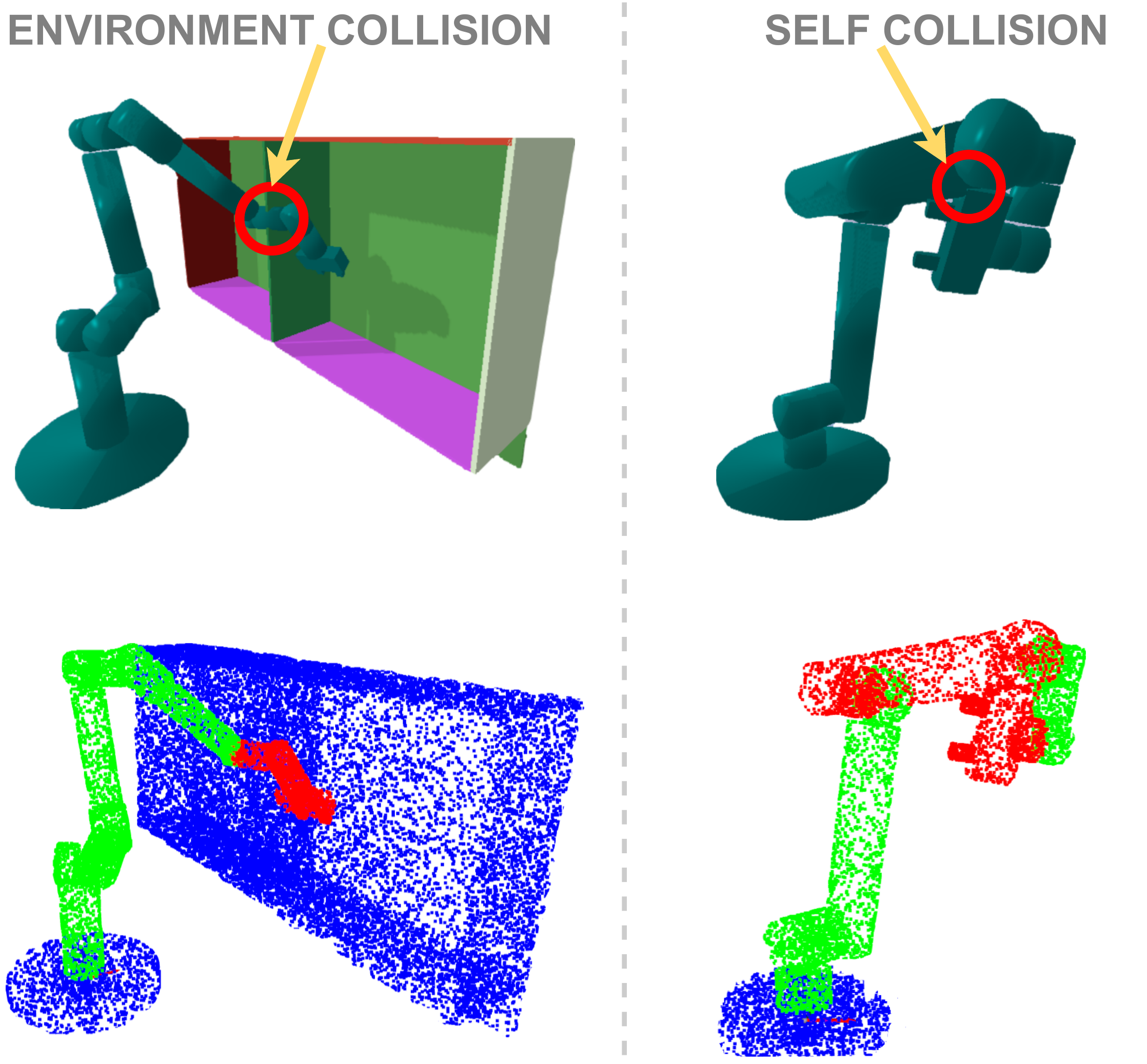}
    \caption{\small Point-wise training labels for XCoD. \includegraphics[height=7.5pt]{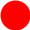} Collision, \includegraphics[height=7.5pt]{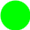} Not Collision, \includegraphics[height=7.5pt]{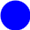} Obstacles.}
    \label{fig:collision_labels}
    \vspace*{-10pt}
\end{figure}
Fig.~\ref{fig:collision_labels} shows planning scenes and corresponding colorized pointclouds highlighting point-wise training labels. We utilize a Point Transformer V3 (PTv3)~\cite{wu2023point} model for the semantic segmentation task and train it using cross-entropy loss along with an additional surrogate Lovász hinge loss, that has shown to improve semantic segmentation performance in prior work~\cite{berman2018lovasz}.

We use XCoD to assign scores for a batch of trajectories predicted by XMoP and choose the future trajectory with the fewest collisions for locally reactive planning. Eq.~\ref{eq:mpc_formula} shows the formulation of the proposed Model Predictive Control (MPC) method where $N$ is the number of surface points sampled from the manipulator, $B$ is the MPC batch size, and $\hat{y}$ is the per-point collision logit from XCoD.
\begin{equation}
    \label{eq:mpc_formula}
    \begin{aligned}
    & a^{*}_t = a^{(q)}_t, \quad q = \argmin [s_1, s_2, \ldots, s_B],  \\
    & s = \frac{1}{HN}\sum_{h=1}^{H}\sum_{i=1}^{N} \argmax \hat{y}_i
    \end{aligned}
\end{equation}
Similar to Diffusion Policy \cite{chi2023diffusion}, we set the prediction horizon $H_p=16$, while the execution horizon $H_a$ is determined on the fly based on the collision scores from XCoD during MPC rollouts. We empirically found that $H_a$ between $2$ to $4$ with MPC batch size of $B=16$ works best for both simulation and real-world experiments.

\subsection{Data Generation and Training}
\label{sec:data_details}
\textbf{Kinematic Templates}: Synthetic manipulators are represented with open kinematic chains connecting a series of rigid-body links. We design these links using $3$ axis-aligned cylinders forming a rigid-body template. Each link template is parameterized with the following information: (1) length of the cylinders, (2) radius of the cylinders, and (3) constraints for the joint that connects the link to the preceding link. We follow the design pattern of two commercially available robots: (1) $6$-DoF UR~\cite{universalrobots} (2) $7$-DoF Sawyer~\cite{rethinkrobotics}. Fig.~\ref{fig:poster} (\textit{left}) shows composed manipulators sampled from our synthetic embodiment distribution by randomizing the parameters for constituent link templates. We adopt the $3.27$ million synthetic planning problems from the M$\pi$Nets dataset \cite{fishman2023motion} and generate demonstration data by sampling a unique embodiment for each problem and solving it using the AIT$^\ast$~\cite{strub2022adaptively} motion planner.

\begin{figure}[h]
    \vspace*{-5pt}
    \centering
    \includegraphics[width=0.33\textwidth]{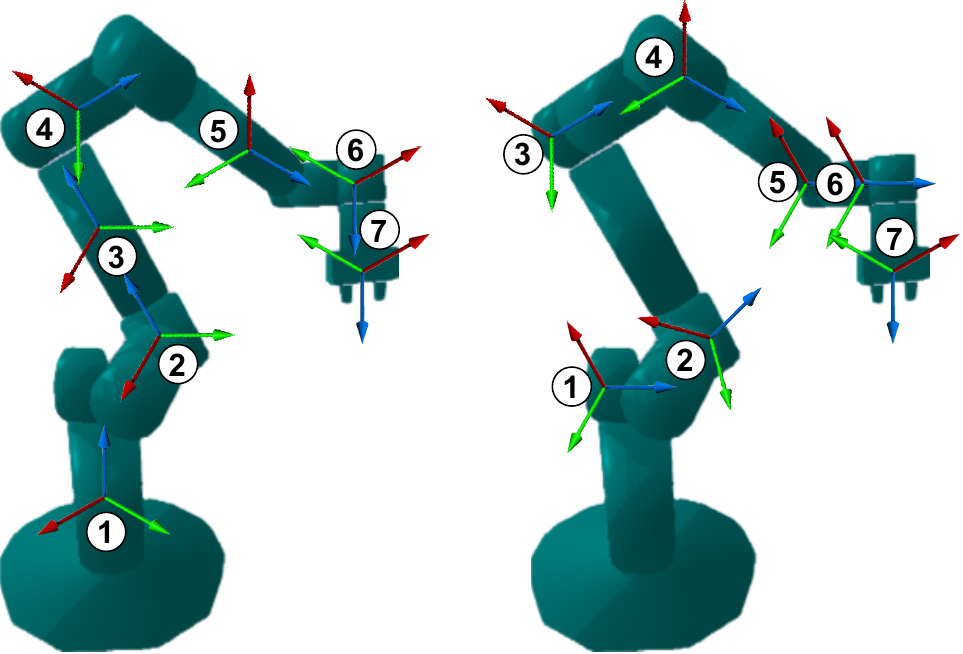}
    \caption{\small Two possible sets of frame sequences for a sampled $7$-DoF manipulator. \includegraphics[height=7.5pt]{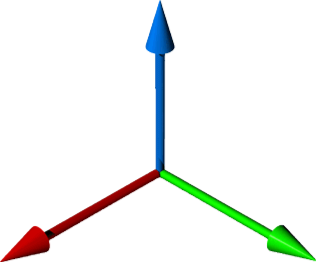} Link frame considered for \textit{pose-token}.} 
    \label{fig:frame_augmentation}
    \vspace*{-10pt}
\end{figure}

\textbf{Data Augmentation}: During training, we randomize the position and orientation of the link frames for pose computation by uniformly choosing cylinders from the constituent link templates. Fig.~\ref{fig:frame_augmentation} shows two possible frame sequences for a sampled robot. With the frame augmentation technique, number of possible sequences for a single manipulator is $3^D$ which promotes cross-embodiment generalization during training. Additionally, we adopt the data augmentation strategies from M$\pi$Nets~\cite{fishman2023motion} and add Gaussian noise to joint and pointcloud observations during training while also sampling unique pointclouds during each training iteration to promote sim-to-real visual robustness.

\section{Experiments and Results}
\label{sec:results}
It should be noted that none of the robots used in our benchmark or real-world experiments were part of XMoP's training dataset. All of our results (simulation and real-world) are \textit{zero-shot} evaluations using a single set of frozen XMoP~($38.7$M) and XCoD~($2.5$M) checkpoints, which were completely trained on synthetic planning demonstration data. 

\subsection{Simulation Benchmark Evaluations}
\label{sec:benchmarks}
\renewcommand{\arraystretch}{1.5} 
\begin{table*}[t]
\centering
\resizebox{0.9\textwidth}{!}{
\begin{tabular}{c | c c c | c c c | c c c }
\toprule
\multirow{2}{*}{\textbf{Embodiment}} & \multicolumn{3}{c|}{\textbf{XMoP+XCoD}} & \multicolumn{3}{c|}{\textbf{AIT$^\ast$+XCoD}} & \multicolumn{3}{c}{\textbf{AIT$^\ast$+PyBullet}} \\ 
 & SR[\%] $\uparrow$ & PL $\downarrow$ & ST[s] $\downarrow$& SR[\%] $\uparrow$ & PL $\downarrow$ & ST[s] $\downarrow$& SR[\%] $\uparrow$ & PL $\downarrow$ & ST[s] $\downarrow$\\
\midrule
Panda  
& 71.8 & 4.6 $\pm$ 4.7 & 49.8 $\pm$ 65.8
& 86.0 & 3.6 $\pm$ 2.3 & 39.7 $\pm$ 27.3
& 94.4 & 2.9 $\pm$ 1.5 & 4.0 $\pm$ 0.3 \\ 
Sawyer  
& 70.8 & 4.8 $\pm$ 5.3 & 42.9 $\pm$ 53.6 
& 90.4 & 3.3 $\pm$ 2.8 & 34.6 $\pm$ 27.1 
& 92.4 & 1.9 $\pm$ 0.9 & 3.9 $\pm$ 0.5 \\ 
IIWA  
& 71.0 & 5.1 $\pm$ 5.6 & 38.3 $\pm$ 52.5 
& 87.6 & 2.8 $\pm$ 2.1 & 32.3 $\pm$ 21.2 
& 93.4 & 2.1 $\pm$ 1.0 & 3.9 $\pm$ 0.4 \\ 
Gen3 $6$-DoF   
& 67.6 & 4.7 $\pm$ 6.0 & 51.8 $\pm$ 70.9 
& 71.0 & 2.5 $\pm$ 2.6 &  24.2 $\pm$ 11.0
& 92.4 & 2.0 $\pm$ 0.9 & 3.9 $\pm$ 0.5 \\ 
Gen3 $7$-DoF  
& 78.2 & 5.5 $\pm$ 5.9 & 44.0 $\pm$ 53.0 
& 88.4 & 3.3 $\pm$ 2.6 & 35.0 $\pm$ 22.6 
& 94.2 & 2.2 $\pm$ 1.2 & 3.9 $\pm$ 0.4 \\ 
UR$5$  
& 70.8 & 3.1 $\pm$ 3.3 & 42.2 $\pm$ 71.2 
& 80.8 & 2.6 $\pm$ 1.9 & 31.0 $\pm$ 20.7 
& 88.8 & 2.1 $\pm$ 1.5 & 3.9 $\pm$ 0.4 \\ 
UR$10$  
& 67.4 & 3.1 $\pm$ 3.4 & 31.5 $\pm$ 52.0 
& 72.6 & 2.9 $\pm$ 2.6 & 33.2 $\pm$ 24.8 
& 92.2 & 2.1 $\pm$ 1.2 & 3.8 $\pm$ 0.6 \\ 
\bottomrule
\end{tabular}%
}
\caption{\small Benchmark results. There are no neural planning baselines that work across multiple embodiments (see \ref{sec:failed_exp}). Hence, we compare to the upper baselines of AIT$^\ast$+PyBullet which we expect to perform better as it has access to the ground truth obstacle geometry information. In contrast, XMoP plans directly from visual inputs. See \ref{sec:result_analysis} for discussion on the hybrid baseline. All inference are on RTX A5000 GPU.}
\label{tab:obs_bench}
\vspace*{-10pt}
\end{table*}

\begin{table}[t]
\centering
\resizebox{\columnwidth}{!}{
\begin{tabular}{c | c c c }
\toprule
\multirow{2}{*}{\textbf{Robot}} & \multicolumn{3}{c}{\textbf{Planning Success Rate ($\%$)}} \\ 
 & Unstructured Obstacles & Wall Hopping & Bin-to-Bin \\
\midrule
Franka FR3  
& 70.0 & 80.0 & 70.0 \\ 
Rethink Sawyer  
& 80.0 & 80.0 & 50.0 \\ 
\bottomrule
\end{tabular}
}
\caption{\small Sim-to-real results on three real-world domains. The average duration of these experiments including planning and rollout was $43.77 \pm 22.20$ seconds. All inference are on RTX 3090 GPU.}
\label{tab:real_world_eval}
\vspace*{-10pt}
\end{table}
We evaluate XMoP on $7$ different robotic manipulators from $5$ commercial manufacturers: Franka Panda, Rethink Sawyer, Kuka IIWA, Kinova Gen3 $6$-DoF, Kinova Gen3 $7$-DoF, Universal Robots UR$5$, and Universal Robots UR$10$. For each manipulator we use a set of $500$ novel problems from the M$\pi$Nets~\cite{fishman2023motion} test distribution, ensuring that valid collision-free IK solutions exist for both start and goal end-effector targets. Policy rollouts are terminated if the manipulator's end-effector reaches the goal or a maximum of $200$ rollout steps are exhausted. We consider a goal to be reached when the $L2$ norm of the $9$D pose (\ref{sec:policy}) difference between end-effector and goal is less than a pre-specified threshold of $0.01$. 

\textbf{Baseline Planners}: We compare our policy against the upper performance threshold of AIT$^\ast$~\cite{strub2022adaptively} planner that has access to an oracle collision checker from PyBullet. We also evaluate generalization capabilities of our learned collision model XCoD by combining it with the AIT$^\ast$ baseline. This hybrid planner utilizes the XCoD model for collision queries.

We utilize the following quantitative metrics to evaluate the planning performance: (1) \textit{Success Rate (SR)} - A trajectory is successful if the final end-effector position is within $1$~cm and orientation is within $5^\circ$ of the goal, with no collisions or joint limit violations. (2) \textit{Path Length (PL)}: Sum of $L2$ norm between consecutive configuration-space way points. (3) \textit{Solution Time (ST)}: Total time elapsed to generate a successful trajectory. Table \ref{tab:obs_bench} shows the benchmark results.

\subsection{Real-world Evaluations}
\label{sec:real_world_eval}
\begin{figure*}[t]
\includegraphics[width=0.9\textwidth]{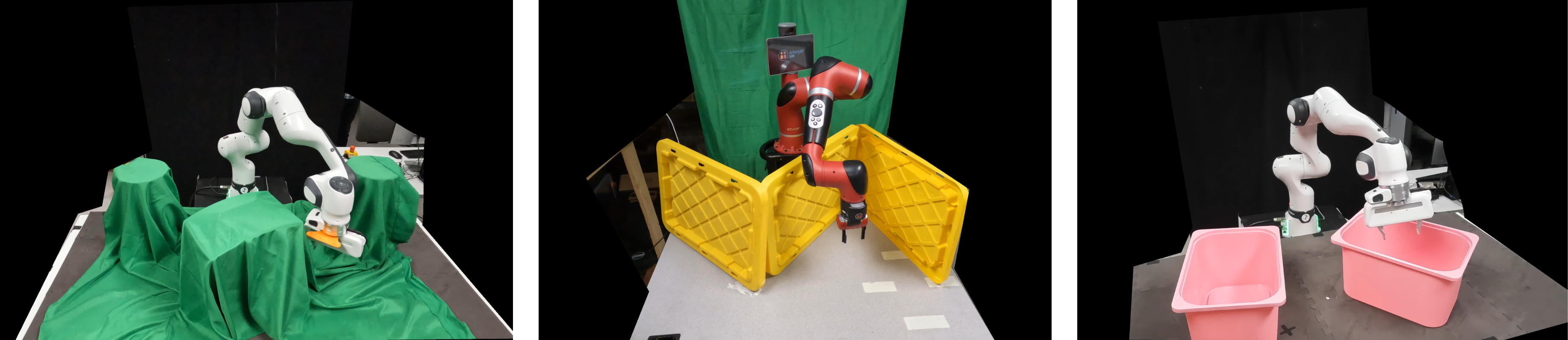}
\centering
\caption{\small XMoP successfully plans for two unseen robotic manipulators in three real-world domains. Unstructured Obstacles (\textit{left}), Wall Hopping (\textit{center}), and Bin-to-Bin (\textit{right}). Videos of policy rollouts are available at \href{https://prabinrath.github.io/xmop}{https://prabinrath.github.io/xmop}.
} 
\label{fig:real_world}
\vspace*{-10pt}
\end{figure*}

Prior works on neural planning do not have any evaluation domains for real-world, hence we created three novel domains for quantitative evaluations and statistics.
\begin{itemize}[leftmargin=*, nolistsep]
    \item \textbf{Unstructured Obstacles Domain}: A domain with random obstacles covered with a green screen to demonstrate planing abilities in  unstructured and cluttered environments.
    \item \textbf{Wall Hopping Domain}: A domain with structured shapes acting as obstacles that occupy a significant amount of space in front of the robot. 
    \item \textbf{Bin-to-Bin Domain}: A task-oriented domain where the robot plans its movement from one bin to another, similar to tasks it might perform in a warehouse environment.
\end{itemize}
Fig.~\ref{fig:real_world} shows the three real-world domains. We used mono-color obstacles and segmented them from a calibrated depth camera to extract the obstacle point cloud. As discussed in \ref{sec:coll_det}, a segmented pointcloud of the robot was augmented into the scene for generating the input for XCoD. We manually assigned end-effector goal poses evenly distributed around the obstacles for each of our experiments. A rollout was considered successful if the manipulator reached the goal without touching any obstacles or itself. All domains had obstacles with dimensions equal to or larger than the links of the manipulator. We conducted 10 experiments for each domain with two different robots Franka FR3 and Rethink Sawyer. XMoP achieved an overall success rate of $71.6\%$. The success rate of our experiments is shown in Table~\ref{tab:real_world_eval}.

\subsection{Result Analysis and Discussion}
\label{sec:result_analysis}
\textbf{Plan Optimality}:
Table \ref{tab:obs_bench} shows that the AIT$^\ast$ upper baseline generates approximately $50\%$ more optimal plans and is $10$ times faster compared to the XMoP planner. However, this comes with the assumption of privileged obstacle information being available for collision detection, thus making it difficult to deploy in unseen real-world environments. We believe our work is a preliminary step in the direction of learning neural planners independent of the embodiment, thus bringing their capabilities one step closer to classical motion planners.

\textbf{Collision Detection}: As shown in Table \ref{tab:obs_bench}, the hybrid AIT$^\ast$+XCoD planner achieves an average success rate of $82.4\%$, which is within $10\%$ of the oracle baseline, demonstrating the effectiveness of our learned collision detection model. However, this hybrid planner fails in the real world, as binary collision queries using XCoD are inaccurate with noisy point clouds. In contrast, XMoP's MPC design allows for more robust real-world deployment. XCoD is a generalized collision model that enables collision-free rollouts for MPC-based policies, such as XMoP. Although XCoD is trained on fully observable synthetic pointclouds, we found that the backbone PTv3~\cite{wu2023point} model is effective in handling partial pointclouds captured from real-world depth cameras. 

\textbf{Computational Efficiency}: The current SOTA neural planner M$\pi$Nets~\cite{fishman2023motion} needs two weeks of training for a single manipulator and requires a compute heavy data generation effort. In contrast, XMoP needs \textit{one-time} data generation and takes two days to train on a single RTX A5000 GPU.

\textbf{\textit{Zero-shot} Generalization}: Table \ref{tab:obs_bench} demonstrates XMoP's ability to plan for manipulators with novel designs that were unseen during training. XMoP exploits the fact that motion behavior is characterized by the whole-body pose of the embodiment~\cite{arduengo2021human, cheng2024expressive}. For similar whole-body poses, different manipulators might have contrasting joint configurations, but the link poses are relatively closer in SE($3$). Similarly, configuration-space actions for different manipulators are dependent on their morphology, but link pose transformations are similar for individual manipulator links. 

\textbf{Sim-to-real Experiments}: We found the Bin-to-Bin domain particularly challenging, as the policy needed to accurately plan the approach angle to avoid collisions with the bin walls (see Table \ref{tab:real_world_eval}). However, this experiment demonstrates a practical use case where XMoP can be applied for \textit{zero-shot} task execution in unseen real-world environments.

\textbf{Kinematic Constraints Satisfaction}: Our policy predicts link-wise transformations that obey kinematics constraints across different manipulators. Our hypothesis is that it learns a correlation between \textit{pose-token} sequence and the distribution of kinematically feasible whole-body transformations. We are actively investigating this property of XMoP and will provide a more detailed analysis in our future work.

\textbf{Limitations}: XMoP's performance is limited by the quality of synthetic training data, leading to struggles with highly out-of-distribution (OOD) planning setups. Moreover, XCoD collision checking is slow, accounting for approximately $90\%$ of the solution time (see Table \ref{tab:obs_bench}), which could be improved by using better semantic segmentation models in the future.

\subsection{Failed Baselines and Ablations}
\label{sec:failed_exp}
\textbf{M$\pi$Nets Baseline}: Table \ref{tab:mpinet_baseline} shows the benchmark results for the M$\pi$Nets model trained on demonstration data from the Franka Panda robot. We also trained a similar model on XMoP training data, which overfitted to the geometric design of synthetic robots and could not transfer \textit{zero-shot} to any of the unseen robotic manipulators (resulting in $0\%$ SR). On the contrary, XMoP achieves an average SR of $71.1\%$ across $7$ unseen robots as shown in Table \ref{tab:obs_bench}. 
\begin{table}[h]
\centering
\resizebox{\columnwidth}{!}{
\begin{tabular}{c | c c c c c c c }
\toprule
\multirow{2}{*}{\textbf{Baseline}} & \multicolumn{7}{c}{\textbf{Planning Success Rate ($\%$)}} \\ 
 & Panda & Sawyer & IIWA & Gen3 6-DoF & Gen3 7-DoF & UR5 & UR10 \\
\midrule
M$\pi$Nets & 89.6 & 0 & 0 & 0 & 0 & 0 & 0 \\
\bottomrule
\end{tabular}
}
\caption{\small The SOTA neural planner M$\pi$Nets pre-trained on single embodiment data fails to generalize across unseen robots.}
\label{tab:mpinet_baseline}
\vspace*{-5pt}
\end{table}

\textbf{ACT Baseline}: We evaluated the SOTA C-space behavior cloning policy ACT~\cite{zhao2023learning} for cross-embodiment planning. This policy was provided with privileged embodiment information including link lengths, link radius, and joint limits. However, it failed to control ($0\%$ SR) both unseen synthetic and commercial robots. In contrast to prior works discussed in \ref{sec:related_work}, XMoP control policies are not trained on any embodiment-specific information, thus showing that such information is not absolutely necessary for cross-embodiment generalization.

\textbf{Ablation Studies}: We ablated the XMoP policy to understand the importance of each of our design decisions.
\begin{itemize}[leftmargin=*, nolistsep]
    \item \textbf{w/o SE($3$) Proprioception}: This is the most critical component for cross-embodiment generalization without which the SR drops to $0\%$.
    \item \textbf{w/o Kinematic Masking and Morphology Adaptation}: We completely removed the masking scheme, thus allowing every \textit{pose-token} to attend every other \textit{pose-token}. Compared to XMoP, the average SR dropped by $4.4\%$.
    \item \textbf{w/o Link Horizon Position Embedding}: We replaced the proposed position embedding scheme with learned position embeddings from the Transformer paper \cite{vaswani2017attention}. The average SR compared to XMoP dropped by $1.2\%$.
    \item \textbf{w/o Frame Augmentation}: We did not use the frame augmentation technique from \ref{sec:data_details} and instead used the pose of the first cylinder in each link template. Without frame augmentation, XMoP's average SR dropped by $68.1\%$.
    \item \textbf{Scale of Data}: We trained XMoP with reduced dataset sizes i.e., $1$M and $2$M demonstrations which resulted in $51.8\%$ and $23.2\%$ drop in average SR respectively.
\end{itemize}
In summary, we observed that the masking scheme and position embedding scheme do not significantly contribute to the success of XMoP. However, SE($3$) proprioception, frame augmentation, and higher scale of data are critical for cross-embodiment generalization.

\section{Conclusion}
\label{sec:conclusion}
In this paper, we presented XMoP, a novel configuration-space neural motion policy that solves novel planning problems \textit{zero-shot} for unseen robotic manipulators which has not been achieved by any prior robot learning algorithm. We formulated C-space control as a link-wise SE($3$) pose transformation method, showcasing its scalability for data-driven policy learning. We used fully synthetic data to train models for motion planning and collision detection while demonstrating strong sim-to-real generalization with a $70\%$ success rate. Our work demonstrates for the first time that C-space motion policies can be learned without embodiment bias and that these learned behaviors can be transferred to novel unseen embodiments in a \textit{zero-shot} manner. We hope our work will enable more generalized robot foundation models that are capable of generating whole-body motion for a diversity of robots. Our work is a preliminary step towards such desirable generalization in learning robot behaviors.

\section*{ACKNOWLEDGMENT}
We would like to thank Anant Sah, Wei Wei Gu, Omkar Patil and Suresh Kondepudi for discussions and help. This material is based upon work supported by the Air Force Office of Scientific Research under award number FA9550-24-1-0239. Any opinions, findings, and conclusions or recommendations expressed in this material are those of the author(s) and do not necessarily reflect the views of the United States Air Force.  

\bibliographystyle{IEEEtran}
\bibliography{references}

\clearpage
\section*{APPENDIX}
\subsection{Embodiment Sampling with Kinematic Templates}
\label{apx:kinematic_templates}

\begin{figure*}[t]
\includegraphics[width=\textwidth]{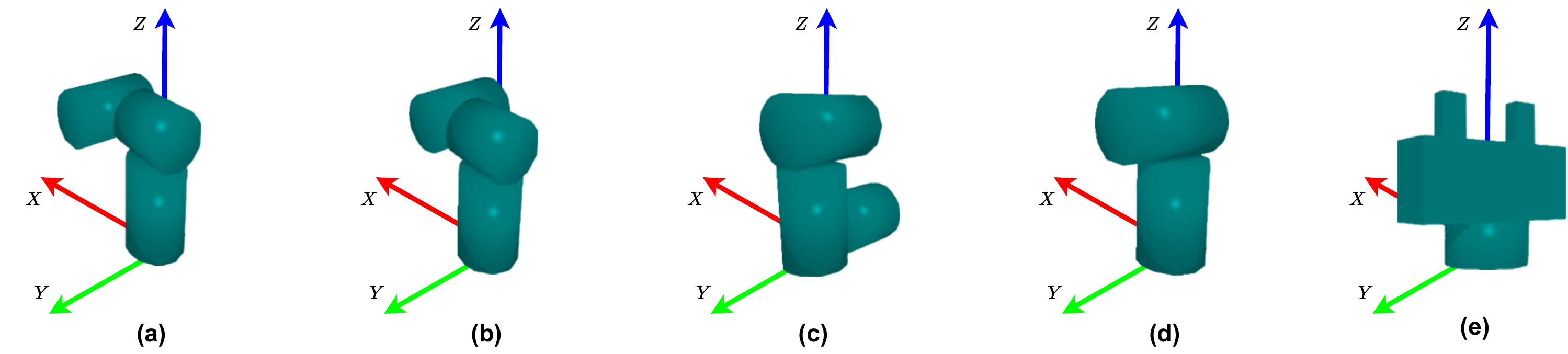}
\centering
\caption{\small Example of random templates that are composed to sample synthetic embodiments for data generation. (a) $3$-Cylinder \texttt{zxy} template with no chiral flip, (b) $3$-Cylinder \texttt{zxy} template with chiral flip, (c) $3$-Cylinder \texttt{yzy} template with chiral flip, (d) $2$-Cylinder \texttt{zxy} template with no chiral flip, (e) End-effector template.} 
\label{fig:kinemtaic_template}
\end{figure*}

\textbf{Link Template}: We design $3$-Cylinder and $2$-Cylinder templates to represent rigid-body links for our synthetic manipulators. These cylinders sequentially extrude along the $3$D axis as shown in Fig.~\ref{fig:kinemtaic_template}(a). Each link is parameterized with six numbers leading to a vector representation for link $i$ as $L_i=(p, l_x, l_y, l_z, r, cf)$ where $i \in \{1,2, \ldots, D\}$ and $D$ is the number of rigid-body links on the manipulator. The definition of the parameters are as follows: 
\begin{itemize}[leftmargin=5mm, nolistsep]
    \item $p$: A sequence of three characters representing connection pattern of cylinders within the link. It has $3^3$ possible permutations i.e., $\{\texttt{xyz}, \texttt{yxz}, \texttt{zxy}, \texttt{yzy}, \ldots\}$. Each pattern maps to a number that is referenced in $L_i$. For example, the pattern \texttt{zxy} means the first cylinder extrudes along $z$ axis, the second cylinder extrudes along $x$ axis, and the third cylinder extrudes along $y$ axis.
    \item $l_x$, $l_y$, $l_z$: Length of the cylinders along $x$, $y$, and $z$ axis respectively.
    \item $r$: Radius of all the three cylinders in the link template.
    \item $cf$: Represents chiral flip \cite{demir2023se} with a binary value of $0$ or $1$. When the value is $1$, a mirror link is created instead of a regular link. Only the last cylinder in the connection pattern is flipped for chirality. A visual representation of a chiral link pair is shown in Fig.~\ref{fig:kinemtaic_template}(a) and Fig.~\ref{fig:kinemtaic_template}(b).    
\end{itemize}
For a $2$-Cylinder template as shown in Fig.~\ref{fig:kinemtaic_template}(d), we set the length of the middle link to 0.

\textbf{End-Effector Template}: The end-effector template is represented with a vector of three numbers $E=(h, r, s, -1, -1, -1)$ where padding $-1$ is used to make the length consistent with the link templates. The end-effector is represented using a $3$-Cuboid mesh on top of a base cylinder as shown in Fig.~\ref{fig:kinemtaic_template}(e). The definition of the parameters are $h$: height of the base cylinder, $r$: radius of the base cylinder, and $s$: scaling factor for the $3$-Cuboid mesh.

\textbf{Joint Constraints}: A pair of link templates $L_{i-1}$ and $L_i$ can be chained together only if the third character in the pattern of link $i-1$ matches with the first character in the pattern of link $i$. Such a chain is connected with a revolute joint parameterized using a vector of two numbers $J_i=(ll, ul)$. The definition of the parameters are $ll$: lower joint limit, and $ul$: upper joint limit.

\begin{figure*}[t]
\includegraphics[width=\textwidth]{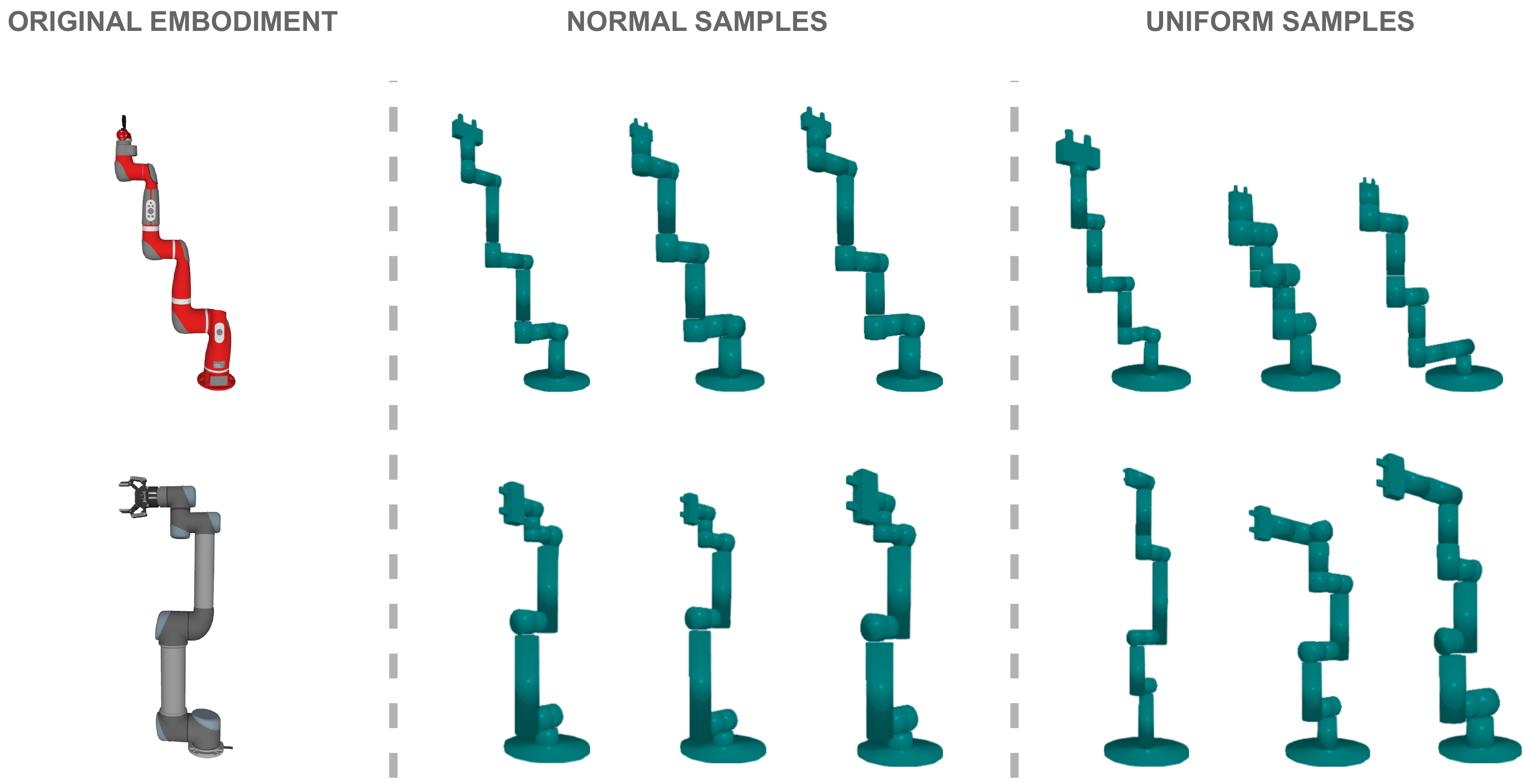}
\centering
\caption{\small Sampled robots from our training distribution. XMoP is trained on planning demonstrations data from such synthetic embodiments and \textit{zero-shot} transfers to unseen manipulators in both simulation and real-world.} 
\label{fig:wall_of_robots}
\end{figure*}

\textbf{Kinematic Template}: We hypothesize that these basic templates can be used to compose arbitrary manipulators that are kinematically valid and lie within the distribution of commercial manipulators. A synthetic manipulator is represented using matrix $\texttt{KT}$ of shape $(8 \times D)$ as shown in eq.~\ref{eq:kinematic_template}.
\begin{equation}
    \texttt{KT} = \begin{bmatrix}
    L_1 & L_2 & \cdots & L_{D-1} & E \\
    J_1 & J_2 & \cdots & J_{D-1} & J_D \\
\end{bmatrix}
\label{eq:kinematic_template}
\end{equation}
We fix the patterns $p$ and chiral flips $cf$ to match with that of the commercial robots Sawyer, and UR. Rest of the parameters in the kinematic template are sampled using one of the following strategies:
\begin{itemize}[leftmargin=5mm, nolistsep]
    \item \texttt{normal}: Parameters are sampled from a normal distribution centered at the estimated parameters from Sawyer, and UR robots. The variance is hard specified for individual parameters to ensure that the generated robots are valid.
    
    \item \texttt{uniform}: Parameters are sampled from a uniform distribution with hard conditions to ensure that the generated robots are valid.
\end{itemize}
Fig.~\ref{fig:wall_of_robots} shows sampled manipulators from the distribution of Sawyer, and UR robots. We use ROS \cite{quigley2009ros} \texttt{xacro} to process kinematic templates and generate URDF~\cite{tola2023understanding} files on the fly for data generation.

\subsection{Synthetic Data Generation}
\label{apx:synthetic_data_gen}
\textbf{Planning Problems}: We adopt planning problems from the M$\pi$Nets~\cite{fishman2023motion} dataset, which contains $3.27$ million motion plans for the Franka Panda~\cite{haddadin2022franka} robot across three different indoor environments. We use the terminal joint states of these motion plans to extract start and goal end-effector poses of the Panda robot using forward kinematics function $\phi$. Then, we attempt collision-free IK to find valid start and goal joint configurations, thereby defining novel planning problems with our synthetic manipulators. Similar to \cite{rakita2021collisionik}, we formulate IK as a quadratic optimization problem as shown in eq.~\ref{eq:optim_ik}.
\begin{equation}
    \begin{aligned}
        & \min_{j} f(j) \quad \text{s.t. } j_L \leq j \leq j_U
    \end{aligned}
    \label{eq:optim_ik}
\end{equation}
where $j$ is the configuration-space candidate for optimization, and $j_L$ and $j_U$ are lower and upper joint limits of the manipulator. We define the cost function $f(j)$ as sum of the following objectives:
\begin{itemize}[leftmargin=5mm, nolistsep]
    \item \textit{Position Cost}: Position $x, x_g \in \mathbb{R}^3$ are current and goal positions of the end-effector with respect to the manipulator's base frame.
    \begin{equation}
        \begin{aligned}
            & x = \phi(j).ee.pos \\
            & c_{pos}(x, x_g) = -\exp\left(-\frac{\|x-x_g\|^2}{0.08}\right) + 25 \ast (\|x-x_g\|)^4
        \end{aligned}
        \label{eq:optim_ik_pos}
    \end{equation}
    \item \textit{Orientation Cost}: Quaternion $q, q_g \in \mathbb{R}^4$ are current and goal orientations of the end-effector with respect to the manipulator's base frame. Where $dist(q,q_g)$ is the absolute quaternion distance \cite{horn1987closed} accounting for the sign ambiguity. 
    \begin{equation}
        \begin{aligned}
            & q = \phi(j).ee.rot \\
            & c_{rot}(q, q_g) = -\exp\left(-\frac{dist(q,q_g)^2}{0.08}\right) + 25 \ast (dist(q,q_g))^4
        \end{aligned}
        \label{eq:optim_ik_ori}
    \end{equation}
    \item \textit{Collision Cost}: We use PyBullet to obtain collision distances from the links of the manipulator within an AABB \cite{cai2014collision} collision radius of $0.1$ m for self-collision and $0.3$ m for environment collision. These distances are then passed into an exponential function, $g(x)$, for computing the net collision cost.

    \begin{equation}
        \begin{aligned}
            & g(x) = 
            \begin{cases} 
                1.8^{(-x + 0.01)} & \text{if } (x-0.01)<0, \\
                0 & \text{otherwise},
            \end{cases} \\
            & c_{coll}(j) = \sum_{x \in \text{AABB}(j)} g(x)
        \end{aligned}
        \label{eq:optim_ik_coll}
    \end{equation}
\end{itemize}
For eq.~\ref{eq:optim_ik_pos} and \ref{eq:optim_ik_ori}, we use the groove objective from \cite{rakita2021collisionik} with default parameters from the paper; whereas for eq.~\ref{eq:optim_ik_coll} we empirically found that the exponential objective provides faster optimization convergence. We use the SLSQP implementation from \texttt{scipy}~\cite{virtanen2020scipy} and attempt IK five times with random resets if the optimizer gets stuck in a local minima. Each optimization attempt is capped to a maximum of $500$ iterations.

\textbf{Planning Data Generation}: We employ the SOTA sampling-based planner AIT$^\ast$ \citep{strub2022adaptively} from OMPL \cite{sucan2012the-open-motion-planning-library} along with PyBullet's oracle collision checker to generate a valid solution trajectory. For demonstration optimality, we enforce a minimum search time of $15$ seconds with a maximum $20$ seconds timeout while optimizing over the path length objective. Additionally we set a maximum allowed path length (\ref{sec:benchmarks}) threshold of $10$ to avoid generating highly non-optimal trajectories. To aid the non-convex optimization as discussed in \ref{sec:method}, we re-time the solution path using more granular collision checks along a B-spline interpolation of the generated trajectory allowing a maximum per joint velocity of $0.05$ rad/s. We resample a new embodiment until we solve a given environment, thus ensuring valid motion plans for all the $3.27$ million problems in M$\pi$Nets dataset. The final motion plan along with the kinematic template $\texttt{KT}$ are saved to the disk for training purposes. Our data generation process for XMoP was executed on a cluster of $150$ cloud compute cores all solving planning problems parallely for two weeks.

\textbf{Collision Data Generation}: For every planning problem in our synthetic dataset, we randomly sample an intermediate joint configuration from the saved motion plan for collision data generation. This intermediate configuration is guaranteed to be collision-free as it is part of a valid motion plan. To introduce collision into the scene, we add scaled Gaussian noise as shown in eq.~\ref{eq:joint_noise} where $j \in \mathbb{R}^\text{DoF}$ is the collision-free joint configuration. The uniform scaling factor $\eta$ ensures that the collision model is trained on various noise levels and remains reactive to subtle collisions of the end-effector typically encountered while moving close to environment obstacles. To prevent class specific bias, we ensure equal number of collision and collision-free joint configurations for training XCoD.
\begin{equation}
    \begin{aligned}
        & j_{\text{noisy}} = j + \eta * \mathcal{N}(0, I), \quad \eta = \text{Uniform}(0, 1) \\
    \end{aligned}
    \label{eq:joint_noise}
\end{equation}
\begin{algorithm}[!t]
\caption{WholeBodyIK}
\label{alg:wbik}
\begin{algorithmic}[1]
\State \textbf{Input:} Manipulator \texttt{URDF}, Whole-body predicted pose $\hat{p}_{t+k}$, Configuration-space proprioception $j_t$, Forward kinematics function $\phi$, Max attempts $M$, Cost threshold $\kappa$
\For {\textit{attempts} $1,2,\ldots,M$}
    \State $obj \gets \|\hat{p}_{t+k}-\phi(\texttt{URDF}, j_{t+k})\|$ \Comment{Set objective}
    \State $bounds \gets \texttt{URDF}$ \Comment{Get joint limits}
    \State $j_{t+k}, cost \gets \texttt{SLSQP.minimize}(obj, bounds, j_t)$ 
    \If {$cost < \kappa$}
        \State \textbf{return} $j_{t+k}$
    \EndIf
    \State $j_t \gets j_t + 0.1*\mathcal{N}(0,I)$ \Comment{Perturb and retry} 
\EndFor
\end{algorithmic}
\end{algorithm}

\subsection{Whole-Body IK for Configuration-Space Retrieval}
\label{apx:joint_state_retrieval}
The exact procedure for whole-body IK is shown in Algorithm \ref{alg:wbik}. We attempt to retrieve the configuration-space candidate $j_{t+k}$ using a quadratic program, starting with the initial guess $j_t$, which is the instantaneous joint state proprioception from the manipulator. We add Gaussian noise to $j_t$ and re-attempt IK for ten times if the optimization gets stuck in a local minima. In practice, we find that $100$ SLSQP iterations are enough for convergence with most manipulators.

\subsection{Implementation Details}
\label{apx:architecture_hyperparameters}
\begin{figure*}[!t]
\includegraphics[width=\textwidth]{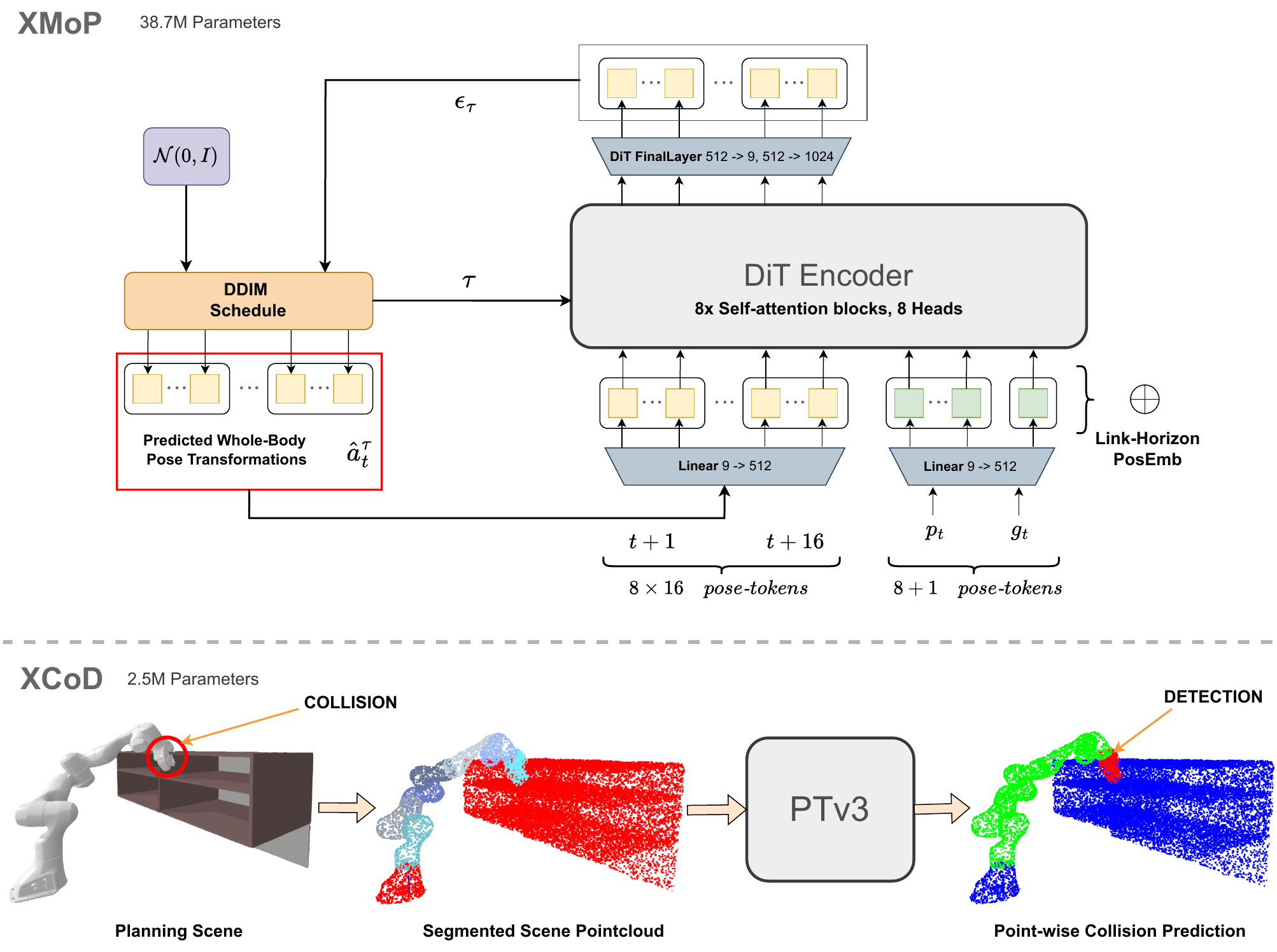}
\centering
\caption{\small Architecture diagram of different components in XMoP. Our planning policy uses self-attention to convert query \textit{pose-tokens} into link-wise relative pose transformations $\hat{a}_t$ conditioned on whole-body poses $p_t$ and goal end-effector pose $g_t$. The collision model XCoD takes a link-wise segmented pointcloud of the planning scene and predicts point-wise collision labels for the manipulator.} 
\label{fig:policy_arch}
\end{figure*}

\textbf{Model Architecture}: A detailed architecture diagram of XMoP is shown in Fig.~\ref{fig:policy_arch}. As discussed in \ref{sec:method} and \ref{sec:policy}, we use $D=8$ (number of rigid-body links) and $H=16$ (prediction horizon) as XMoP parameters. This makes the input sequence length as $D\times(H+1)+1 = 137$ \textit{pose-tokens}.

\begin{figure}[t]
    \centering    
    \includegraphics[width=0.4\textwidth]{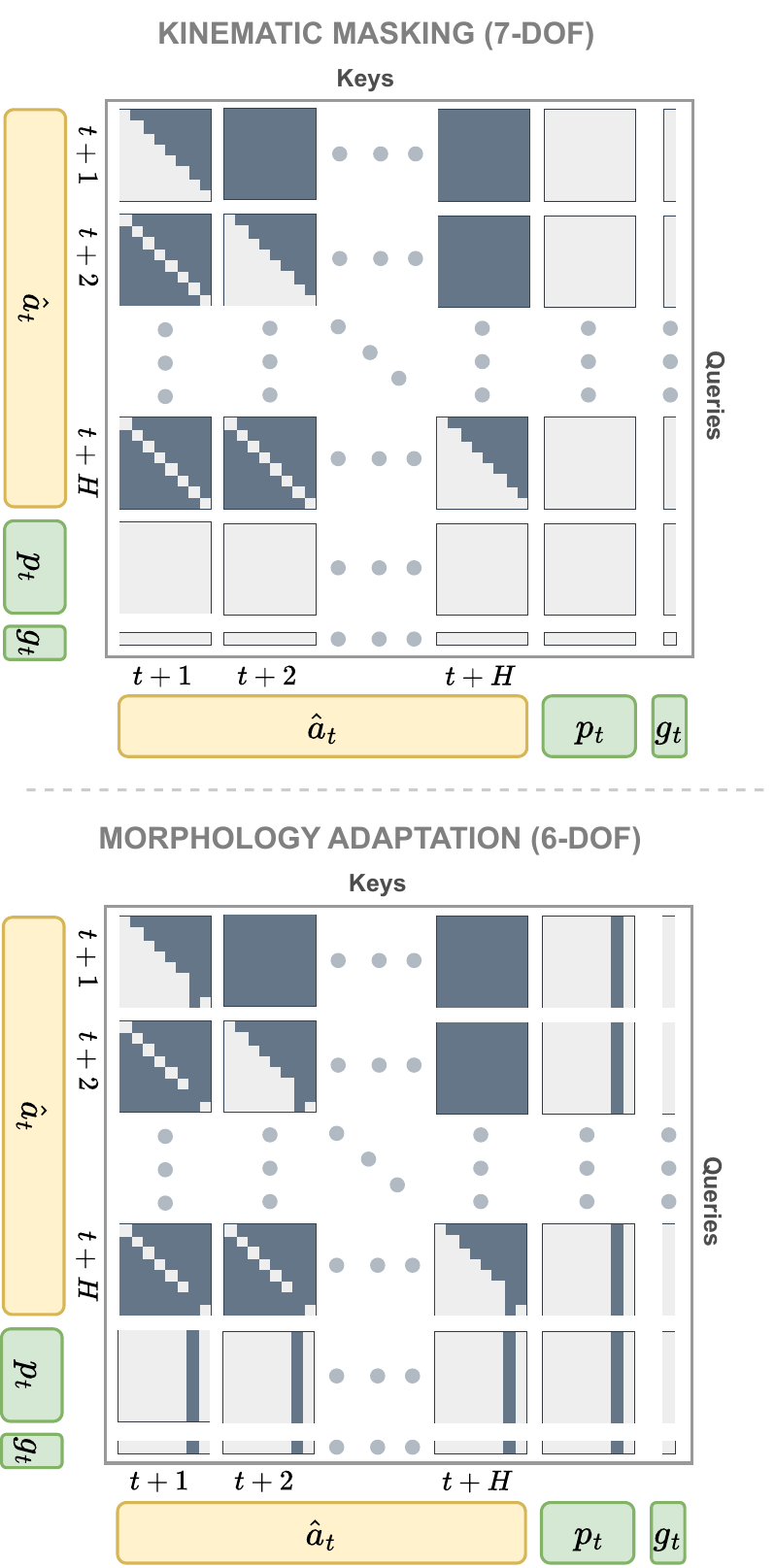}
    \caption{\small Self-attention masking in XMoP. \includegraphics[height=7.5pt]{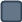} Masked, \includegraphics[height=7.5pt]{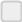} Unmasked, \includegraphics[height=7.5pt]{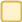} Query \textit{pose-tokens}, \includegraphics[height=7.5pt]{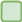} Condition \textit{pose-tokens}. Kinematic mask restricts the self-attention to parent or ancestor links within a horizon step, and to the same link in current and previous horizon steps. Morphology adaption for $6$-DoF is enabled by masking out keys of the unused \textit{pose-token}.} 
    \label{fig:pose_tokens}
\end{figure}

\textbf{Masking Strategy}: As discussed in \ref{sec:policy}, we restrict self-attention in XMoP using kinematic and morphology masking. Fig.~\ref{fig:pose_tokens} shows the horizon wise breakup of the square attention mask used in XMoP. The $1^{st}$ and $8^{th}$ \textit{pose-tokens} at every horizon step is used for base link and end-effector respectively. However, the 6-DoF robots do not have a value for the $7^{th}$ \textit{pose-token}; therefore, we mask out this key at all horizon steps encouraging morphology adaptation.

\textbf{Loss Masking}: While training XMoP,  we backpropagate the loss only for unmasked morphology \textit{pose-tokens}; i.e, the loss for $7^{th}$ \textit{pose-token} at every horizon step is ignored for $6$-DoF robots. For XCoD, the loss is backpropagated only for the augmented points that were sampled from the surface of the robot.

\textbf{Diffusion Model}: For the diffusion model, we use the DDIM~\cite{song2020denoising} approach with $100$ training iterations and $10$ inference iterations. We use the Square Cosine Schedule~\cite{nichol2021improved} for diffusion noise parameterization which has been shown effective for policy training in prior work~\cite{chi2023diffusion}. Furthermore, we follow the conventional practice of using an Exponential Moving Average (EMA)~\cite{he2020momentum} during training and update the model weights with a decay factor of $0.9999$~\cite{chi2023diffusion, peebles2023scalable}.

\textbf{Training and Inference}: The training and inference algorithms for XMoP control policy is shown in Algorithm \ref{alg:xmop_training} and Algorithm \ref{alg:xmop_inference} respectively. Table \ref{tab:impl_details} shows the training hyperparameters.

\begin{table}[!t]
\centering
\resizebox{\columnwidth}{!}{  
\begin{tabular}{c | c | c}
  \toprule
  & \textbf{XMoP} & \textbf{XCoD} \\ 
  \midrule
  \textit{Optimizer} & AdamW & AdamW \\ 
  \textit{Weight Decay} & 0.05 & 0.05 \\ 
  \textit{Learning Rate} & 1e-4 & 5e-4 \\ 
  \textit{LR Schedule} & Linear to 1e-5 over 1 epoch & Cosine Annealing to 5e-5 \\ 
  \textit{Batch Size} & 64 & 12 \\ 
  \textit{Dataset Size} & $3.27\times 10^6$ & $1\times 10^6$ \\  
  \textit{Epochs} & 20 & 1 \\ 
  \textit{Training Time} & 1 day 22 hours & 5 hours \\ 
  \bottomrule
\end{tabular}
}
\vspace{3pt}
\caption{\small Training hyperparameters for XMoP planner.}
\label{tab:impl_details}
\end{table}

\textbf{Classical Planner Baselines}: We set a planning timeout of $20$ seconds for AIT$^\ast$+PyBullet, and $100$ seconds for AIT$^\ast$+XCoD, with a preferred minimum search time of $4$ seconds and $20$ seconds respectively. For fair comparison, we evaluate the smooth trajectory obtained after path shortcutting and interpolation, as these are necessary for practical deployment of classical planners.

\begin{figure}[!h]
    \centering    
    \includegraphics[width=\columnwidth]{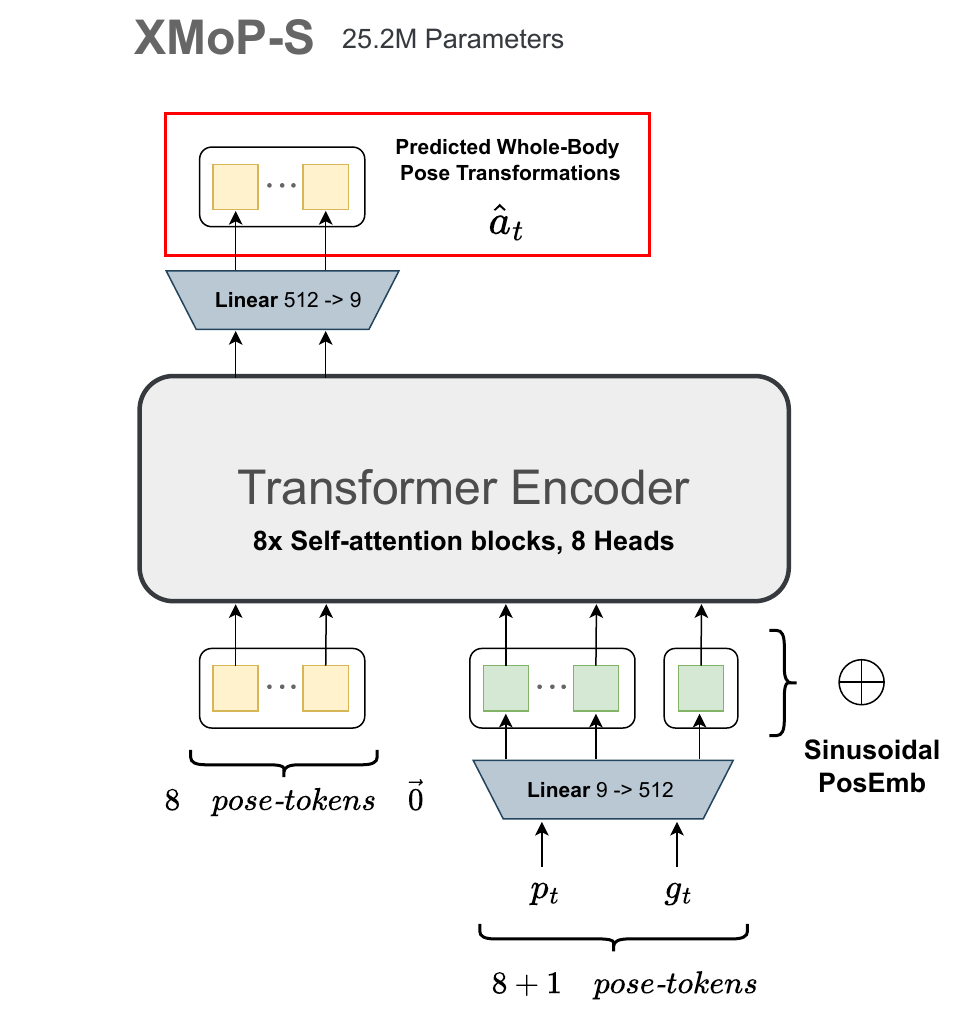}
    \caption{XMoP-S is a deterministic $6$-DoF reaching policy that \textit{zero-shot} generalizes to unseen robotic manipulators.} 
    \label{fig:xmop_s}
\end{figure}

\subsection{XCoD Benchmark Experiments}
\label{apx:additional_benchmarks}
Table \ref{tab:xcod_metrics} shows benchmark evaluations for the learnt collision model XCoD. We evaluate these metrics by sampling equal number of collision and collision-free states. For collision detection we use a binary condition i.e., a configuration is considered to be in collision if the ratio of detected link points in collision to the total number of manipulator points is greater than $0.001$. This condition is also used for AIT$^\ast$+XCoD baseline experiments as shown in Table \ref{tab:obs_bench}. 

\begin{figure*}[t]
\includegraphics[width=\textwidth]{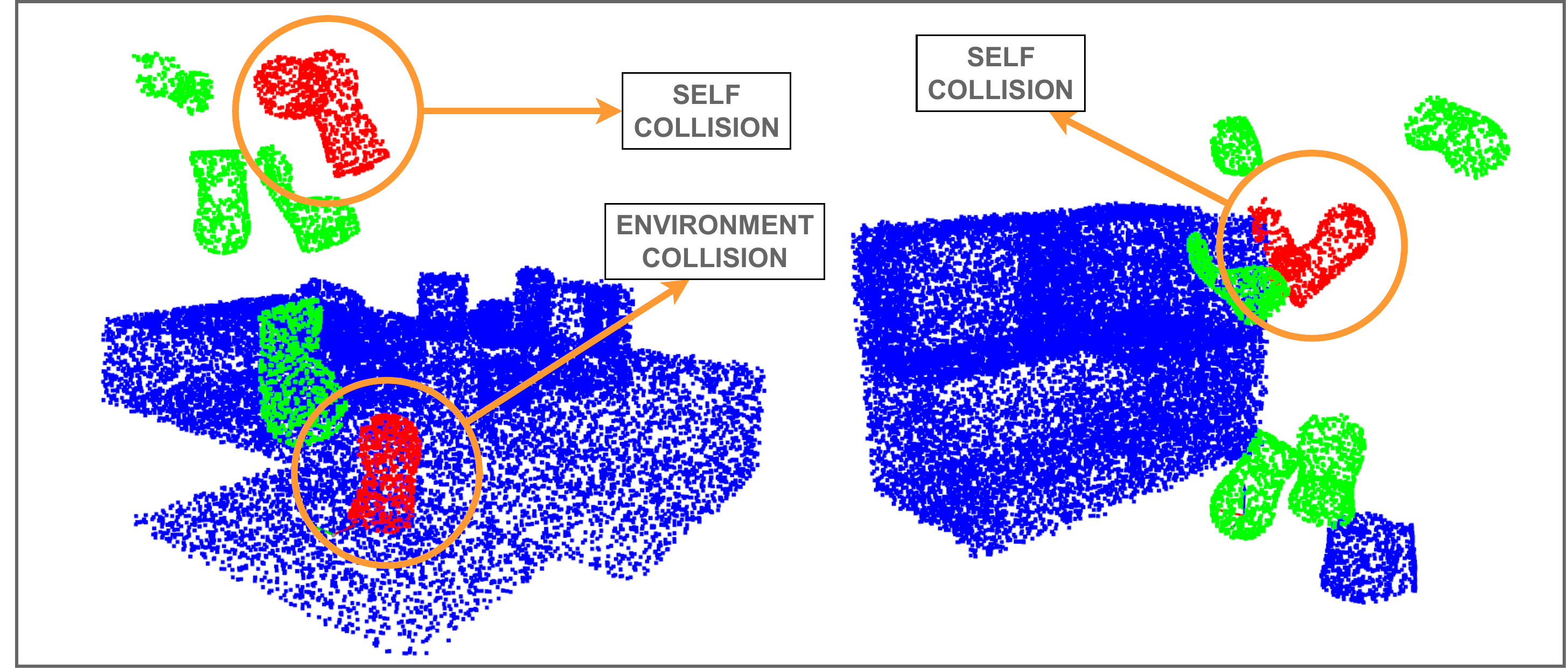}
\centering
\caption{\small Pointcloud captured from a planning scene where the links of Kuka IIWA robot are randomly scattered in 3D space. XCoD identifies self-collisions and environment collisions showcasing highly out-of-distribution generalization for collision detection.} 
\label{fig:causal}
\end{figure*}

The XMoP control policy predicts whole-body poses for future time steps, which are evaluated by XCoD during MPC rollouts. In some out-of-distribution (OOD) scenarios, XMoP's predictions may lack accuracy, leading to spatial distortions in the reconstructed future poses that violate the manipulator's kinematic constraints. To examine XCoD's behavior in such cases, we pushed the collision model to its limits by analyzing highly OOD problems. Collisions occur when the robot makes contact with an environmental obstacle or with itself. Fig.~\ref{fig:causal} presents results from our OOD evaluations, where the links of a Kuka IIWA robot are randomly scattered in 3D space. Our model successfully identifies both self-collisions and collisions with the environment, demonstrating strong generalization for collision detection in highly OOD scenarios.

\begin{algorithm*}[!h]
\caption{Training XMoP control policy}
\label{alg:xmop_training}
\begin{algorithmic}[1]
\State \textbf{Input:} Planning demonstration dataset $\mathcal{D}$, Neural policy $\pi_\theta$, Horizon length $H$, Configuration-space observation noise $\eta$, Forward kinematics function $\phi$
\For {\textit{every training step}} 
    \State Sample $\texttt{KT}$, $\xi$ $\gets$ $\mathcal{D}$ \Comment{Get kinematic template and demonstration trajectory}
    \State $\texttt{URDF} \gets \text{GenerateURDF}(\texttt{KT})$ \Comment{Generate URDF from kinematic template}
    \State Sample $j_{t:t+H}, j_{goal} \gets \xi$ \Comment{Get a trajectory chunk and goal joint configuration}
    \State $j_t \gets j_t + \eta * \mathcal{N}(0, I)$ \Comment{Add C-space observation noise}
    \State $p_t \gets \phi(\sim\texttt{URDF}, j_t)$ \Comment{Get frame randomized whole-body observation pose}
    \State $g_t \gets \phi(\texttt{URDF}, j_{goal}).ee$ \Comment{Get goal end-effector pose}
    \State $p_{t+1:t+H} \gets \phi(\texttt{URDF}, j_{t+1:t+H})$ \Comment{Get target whole-body poses}
    \State $T_{t+1:t+H} \gets p_{t+1:t+H}(p_t)^{-1}$ \Comment{Get whole-body relative pose transformations}
    \State Sample $\tau \sim \text{Uniform}(1, 100)$, $\epsilon \sim \mathcal{N}(0, I)$ \Comment{Get diffusion step and Gaussian noise}
    \State $a^{\tau}_t \gets \text{AddNoise}(T_{t+1:t+H}, \epsilon, \tau)$ \Comment{Add noise using a scheduler}
    \State $\epsilon_\theta \gets \pi_\theta$ \Comment{Get the noise prediction model from policy}
    \State Predict $\hat{\epsilon} \gets \epsilon_\theta(a^{\tau}_t, p_t, g_t, \tau)$ \Comment{Predict diffusion noise}
    \State $\mathcal{L}_{xmop} \gets \text{MSE}(\hat{\epsilon}, \epsilon)$ \Comment{Compute loss}
    \State Update $\theta$ with AdamW and $\mathcal{L}_{xmop}$ \Comment{Update policy parameters with gradient descent}
\EndFor
\end{algorithmic}
\end{algorithm*}

\begin{algorithm*}[!h]
\caption{Inference with XMoP control policy}
\label{alg:xmop_inference}
\begin{algorithmic}[1]
\State \textbf{Input:} Manipulator \texttt{URDF}, Trained neural policy $\pi_\theta$, Configuration-space proprioception $j_t$, Prediction horizon $H$, Goal end-effector pose $g_t$, Forward kinematics function $\phi$, Denoising steps $K$
\State $p_t \gets \phi(\texttt{URDF}, j_t)$ \Comment{Get whole-body observation pose}
\State Sample $\hat{a}^{K}_t \sim \mathcal{N}(0, I)$ \Comment{Get Gaussian noise}
\State $\epsilon_\theta \gets \pi_\theta$ \Comment{Get the noise prediction model from policy}
\For {$\tau=K,K-1,\ldots,1$}
    \State Predict $\hat{\epsilon} \gets \epsilon_\theta(\hat{a}^{\tau}_t, p_t, g_t, \tau)$ \Comment{Predict diffusion noise}
    \State $\hat{a}^{\tau-1}_t \gets \text{Denoise}(\hat{a}^{\tau}_t, \hat{\epsilon}, \tau)$ \Comment{Remove noise using a scheduler}
\EndFor
\State $\hat{p}_{t+1:t+H} \gets \hat{a}^{0}_t p_t$ \Comment{Get predicted whole-body poses}
\For {$k=1,2,\ldots,H$}
    \State $j_{t+k} \gets \text{WholeBodyIK}(\hat{p}_{t+k}, j_t, \texttt{URDF})$ \Comment{Retrieve configuration-space actions}
\EndFor
\end{algorithmic}
\vspace{10pt}
\end{algorithm*}

\begin{table*}[!t]
\centering
\resizebox{\textwidth}{!}{
\begin{tabular}{c|c|c|c|c}
  \toprule
  Embodiment & Segmentation IoU (\%) & Collision Precision (\%) & Collision Recall (\%) & Collision Accuracy (\%) \\ 
  \midrule
  Panda & 79.4 & 96.9 & 98.6 & 97.7 \\ 
  Sawyer & 88.4 & 97.9 & 99.1 & 98.5 \\ 
  IIWA & 85.0 & 99.5 & 99.0 & 99.2  \\ 
  Gen3 $6$-DoF & 67.0 & 71.1 & 99.4 & 79.6 \\ 
  Gen3 $7$-DoF & 67.3 & 81.7 & 98.5 & 88.2 \\ 
  UR$5$  & 73.0 & 97.1 & 66.8 & 82.4 \\ 
  UR$10$  & 76.1 & 98.2 & 70.7 & 84.7 \\ 
  \bottomrule
\end{tabular}
}
\vspace{3pt}
\caption{\small Benchmark results for the learned collision model XCoD. We evaluate the semantic segmentation performance using the Intersection Over Union (IoU) metric. Furthermore, we show the collision detection performance with precision, recall, and accuracy metrics. \textit{Note}: The recall for UR robot was affected by the imprecise collision mesh available for ground-truth validation, which led to false positive labels during evaluations.}
\label{tab:xcod_metrics}
\end{table*}

\subsection{Real-World Experiments}
\label{apx:real_world_experiments}
\begin{figure*}[t]
\includegraphics[width=\textwidth]{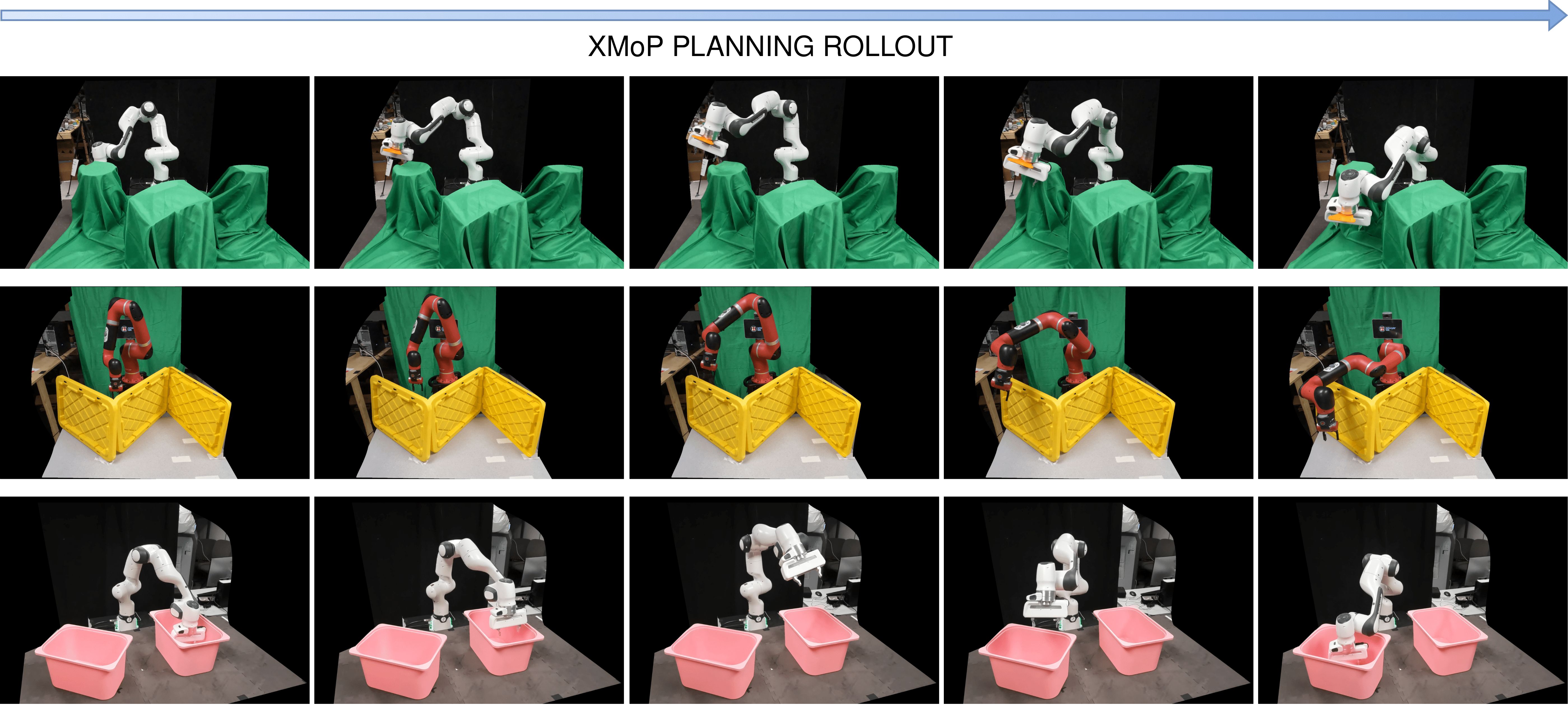}
\centering
\caption{\small XMoP planning rollouts across three unseen real-world domains for two unseen $7$-DoF commercial manipulators Franka FR3 and Sawyer (better viewed when zoomed in). Videos of policy rollouts are available at \href{https://prabinrath.github.io/xmop}{https://prabinrath.github.io/xmop}.} 
\label{fig:planning_snaps}
\end{figure*}

We developed a ROS~\cite{quigley2009ros} RViz-based interface to visualize the environment pointcloud and specify goal end-effector positions for XMoP. For open-loop rollouts, we take a single pointcloud of the obstacles at the beginning and internally rollout the MPC policy to generate the motion plan that is executed on the real robot. Whereas, for closed-loop rollouts, we deploy XMoP policy in real-time. Fig.~\ref{fig:planning_snaps} shows intermediate snapshots of two commercial robots using XMoP to plan across the three real-world domains as discussed in \ref{sec:real_world_eval}.

\subsection{Architecture Ablations}
\label{apx:ablations}
\textbf{Policy Architecture}: We evaluated our whole-body pose transformation method using a comparatively simpler Transformer \cite{vaswani2017attention} policy, as depicted in Fig.~\ref{fig:xmop_s}. In contrast to the original stochastic diffusion policy used in XMoP, this policy was designed as a deterministic variant for predicting whole-body poses over a single horizon step. We refer to this policy as XMoP-S. Due to its deterministic nature, XMoP-S cannot be used with the MPC framework and therefore cannot plan in environments with obstacles. Hence, we assessed its performance on more relaxed $6$-DoF reaching problems by removing obstacles from our benchmark. The XMoP-S policy achieved a $70.9\%$ success rate on these reaching benchmark problems, showcasing that our whole-body control method is applicable to different base control models other than diffusion policies. XMoP-S was trained within $6$ hours on a single A5000 GPU and the policy exhibited real-time performance while operating at an average rollout frequency of $10$ hz.

\textbf{Segmentation Architecture} For our collision model, we also experimented with semantic segmentation using a PointNet++~\cite{qi2017pointnet++} backbone. We found the segmentation performance of PointNet++ to be worse than that of the Point Transformer model \cite{wu2023point}. Furthermore, due to its high memory requirements and latency, we found PointNet++ to be practically infeasible for MPC formulation with XMoP.

\subsection{Failed Ideas}
We tried to integrate a few standard ideas from the neural planning literature, however they performed relatively poorly compared to XMoP:
\begin{itemize}[leftmargin=5mm, nolistsep]
    \item \textbf{C-space Goals for Planning}: When provided with the whole-body pose target as additional \textit{pose-tokens} for goal conditioning, the policy overfits to the sub-optimal planning behaviors of sampling based planners and generates trajectories with higher path lengths.
    \item \textbf{Input Normalization}: Normalization of neural network inputs is a conventional approach that has been used for improving policy performance in prior works \cite{chi2023diffusion, fishman2023motion, zhao2023learning}. However, we did not find it useful for our pose transformation method. In contrast, the input-normalized policy takes longer time to train and fails to generalize to unseen embodiments. 
\end{itemize}

\subsection{Limitations and Future Work}
\label{apx:limitations}

\begin{figure*}[!t]
\includegraphics[width=\textwidth]{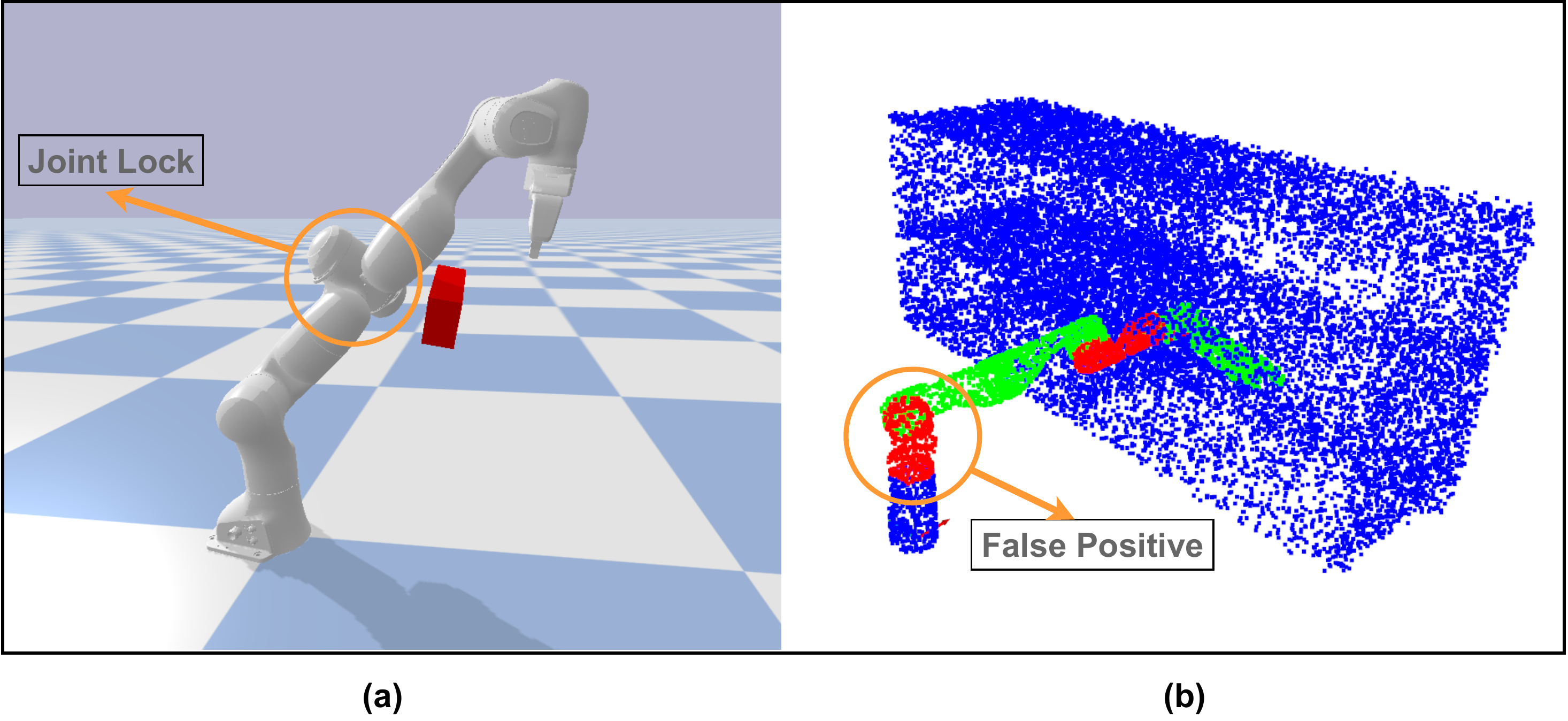}
\centering
\caption{\small Limitations of XMoP planner. (a) As our policy is not conditioned on joint bounds, it runs into joint limits for complex planning tasks. The problem is illustrated for a Panda robot, where our policy does not recover from the locked configuration and fails to plan for the goal pose specified in red. (b) False positive collision detection for a Gen3 $7$-DoF robot.} 
\label{fig:limitations}
\end{figure*}

\textbf{Planning Failures} Our policy struggles to avoid obstacles when the robot's end-effector is close to the specified goal pose. In such scenarios, the generated trajectory samples from the diffusion policy lack the required diversity to avoid obstacles and are too biased toward reaching the goal. Hence, the predicted trajectory with fewest collisions is still a trajectory in collision. A possible solution to this issue could be to add sub-optimal evasive trajectories to the demonstration dataset for near-goal collision avoidance. Moreover, due to the lack of joint bound awareness, for complex planning problems some manipulators get locked at joint limits, as shown in Fig.~\ref{fig:limitations}(a). Further fine-tuning XMoP policy with few demonstrations from the robot might help avoid such undesired behaviors.

\textbf{Collision Failures} Fig.~\ref{fig:limitations}(b) shows an instance of false positive detection from XCoD model when in reality the highlighted robot link is collision-free. We found such instances to be rare and attribute them to the model overfitting to biases in the training data. Moreover, the solution time of our MPC policy is highly dependent on the inference speed of the collision model. Our current implementation allows $1.25$ seconds for each MPC evaluation, which checks $B=16$ different possible future trajectories of the manipulator as predicted by the control policy. We hope future improvements in 3D semantic segmentation methods will enhance the efficiency of XCoD collision detection, thereby improving the inference speed of our MPC policy.

\vspace{10pt}
\textbf{Out-of-Distribution Planning Problems} XMoP policy struggles to plan for OOD goal poses and environment setups, a common issue with behavior cloning methods. Training XMoP on a more diverse planning dataset, where the goals are evenly distributed within the reachable workspace, might help in learning better policies for OOD planning generalization.

\end{document}